\documentclass[conference]{IEEEtran}
\IEEEoverridecommandlockouts

\usepackage{cite}
\usepackage{amsmath,amssymb,amsfonts}
\usepackage{algorithmic}
\usepackage{graphicx}
\usepackage{textcomp}
\usepackage{xcolor}
\usepackage{hyperref}
\usepackage{rotating}

\def\BibTeX{{\rm B\kern-.05em{\sc i\kern-.025em b}\kern-.08em
    T\kern-.1667em\lower.7ex\hbox{E}\kern-.125emX}}

\begin{document}

\title{Towards Self-Supervised High Level Sensor Fusion}

\author{\IEEEauthorblockN{Qadeer Khan}
\IEEEauthorblockA{\textit{TUM and Artisense}}
\and
\IEEEauthorblockN{Torsten Sch\"on}
\IEEEauthorblockA{\textit{Audi Electronics Venture}}
\and
\IEEEauthorblockN{Patrick Wenzel}
\IEEEauthorblockA{\textit{TUM and Artisense}}
}

\maketitle

\begin{abstract}
    In this paper, we present a framework to control a self-driving car by fusing raw information from RGB images and depth maps. A deep neural network architecture is used for mapping the vision and depth information, respectively, to steering commands. This fusion of information from two sensor sources allows to provide redundancy and fault tolerance in the presence of sensor failures. Even if one of the input sensors fails to produce the correct output, the other functioning sensor would still be able to maneuver the car. Such redundancy is crucial in the critical application of self-driving cars. The experimental results have showed that our method is capable of learning to use the relevant sensor information even when one of the sensors fail without any explicit signal.
\end{abstract}

\section{Introduction}\label{chapter:intro}

The fusion of different sensor modalities in the context of autonomous driving is a crucial aspect in order to be robust against sensor failures.  

We consider RGB and Depth camera sensors for training a control module to maneuver a self-driving car.  Depth images have several advantages in that they can be used for distance measurement, object detection, and even determination of vehicle trajectories \cite{SazaraIV2017}. They can also be used to further improve segmentation \cite{HazirbasACCV2016}. However, unlike RGB images, depth images cannot determine the texture, color, brightness etc. of the concerned objects. Color is important in for example determining the status of a traffic light. We would therefore like to utilize the advantages of both sensors to further stabilize vehicle control. The control model trained by fusing of information from the 2 sensors (RGB and Depth) should ensure fault tolerance. Even if one of the 2 sensor fails, the control model should still be able to maneuver the car from the other functioning sensors, thus providing redundancy. Both fault tolerance and redundancy are important characteristics in self driving with multi-sensor input. It is important to ascertain the certainty of predication associated with each of the functioning sensors. \cite{teslacrash2017} is an example of how the car's camera mistook a trailer for the brightly lit sky, resulting in an accident.

For both depth and RGB images we take advantage of the modular approach \cite{WenzelCoRL2018}, where corresponding extracted features are trained separately from the final control module.  We pass the depth image through a pretrained CNN to extract out the depth features as depicted in Figure \ref{fig:sensorfusion}. Meanwhile a separate semantic segmentation network is trained using the RGB images as input. The segmentation network is trained as an encoder-decoder architecture without skip connections. The encoder output produces a latent semantic vector which is a lower dimensional representation of the semantic segmentation output. This latent semantic vector is then concatenated with the depth features to furnish a fused vector which is passed through a control module to give steering commands.  The method is explained in further detail in Section \ref{section:method_sfusion}.  When training the control model by fusing information from these sensors, two things should be considered:

\begin{itemize}
  \item Addition of a sensor gives redundancy, such that if one sensor fails, information from the other sensor still gives the correct steering commands. Without this redundancy, failure of one sensor, may render information from the other functioning sensor to also be useless, since the control module was never trained to predict on one sensor only.
  \item Identify the certainty with which information from sensors can be trusted to perform the correct steering decision.
\end{itemize}

\section{Related Work}\label{chapter:relatedWork}

\noindent{\textbf {Sensor fusion.}} The authors of \cite{PatelIROS2017} demonstrated the fusion of raw information from RGB cameras with depth data obtained from LiDAR for autonomously navigating an indoor robot. Meanwhile,~\cite{KendallNIPS2017} elucidated the importance of knowing the (un)certainty in prediction associated with the sensors concerned. The uncertainties were quantified using Bayesian deep learning methods. These methods provide a probabilistic interpretation of the predictions instead of point estimates.

\section{Method}\label{section:method_sfusion}

\begin{figure}[ht]
  \centering
  \includegraphics[width=\linewidth]{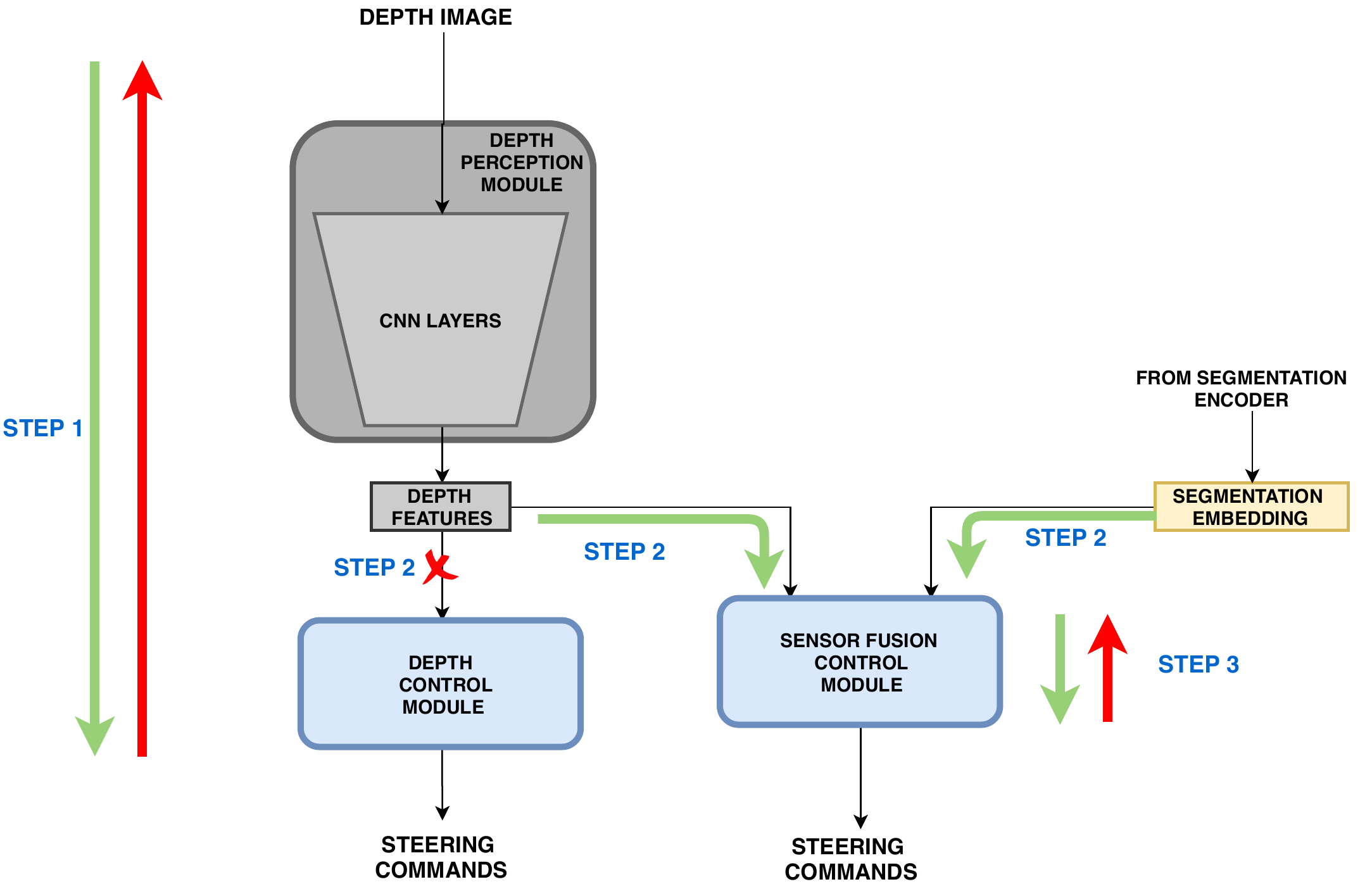}
  \caption{This figure shows how to fuse information from depth maps and semantic embeddings for the purpose of robust car steering control. Black arrows show connections between the various modules. Green and red arrows indicate forward and back-propagation, respectively. \textbf{Step~1:} A model is trained for predicting the steering commands solely from the depth maps. This model is divided into a perception and control module. The perception module learns the depth features which are then passed through the depth control module to predict the steering commands. \textbf{Step~2:} The connection between the depth features and the depth control module is detached. Instead, the depth features along with the semantic latent embeddings are passed through the sensor fusion control module. \textbf{Step~3:} The sensor fusion control module is trained with inputs from the two sensors. Note that in this step, the weights of the depth perception module are fixed.}
  \label{fig:sensorfusion}
\end{figure}

Figure~\ref{fig:sensorfusion} shows the necessary steps for training the (sensor fusion) control module fusing information from the two sensor modalities. Note that the input to the (sensor fusion) control module is a vector concatenated with the segmentation embedding and the depth features. We retrieve the segmentation embedding by training the segmentation network as an encoder-decoder architecture without skip connections~\cite{LarsenICML2016}. The RGB image input to the encoder is of size $128 \times 128$. The encoder output produces a latent semantic embedding (of size $D=64$), which is a lower dimensional representation of the semantic segmentation output. To get the depth features we first independently train a (depth) control model from scratch with only the depth maps of size $128 \times 128$. 

\begin{figure}[ht]
  \centering
  \includegraphics[width=\linewidth]{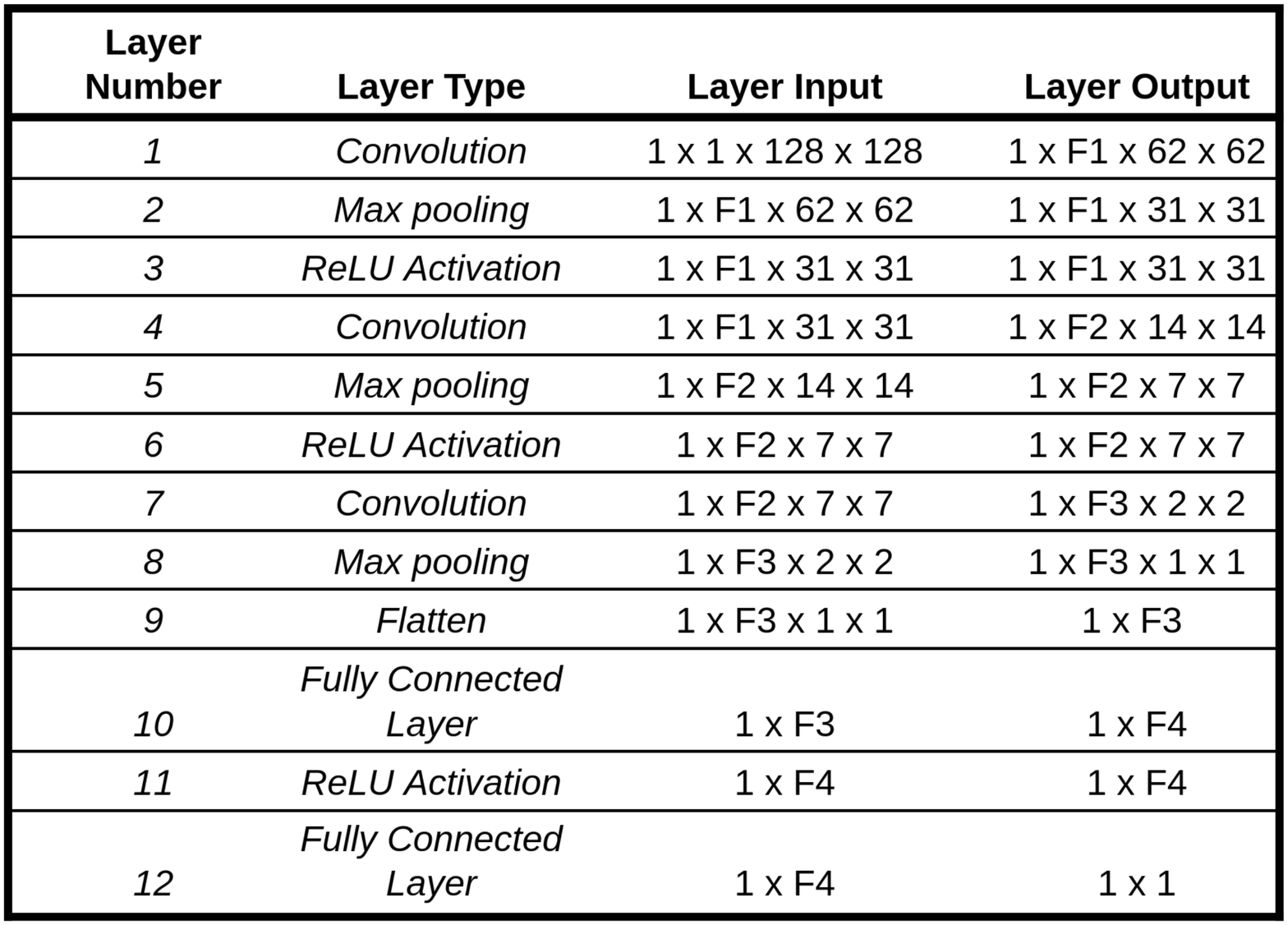}
  \caption{The architecture for training a model controlling the steering command of a car using depth maps. The model can be subdivided into a perception module, which extracts the depth features (layers 1 to 8). These depth features are then flattened (layer 9). The steering angle is predicted by layers 10 to 12 using the depth features as input. $F_1$, $F_2$ and $F_3$ are the number of filters in the convolution represented by layers 1, 4 and 7, respectively. $F_4$ is the number of neurons in the fully connected layer 10. All convolutional layers have a kernel size of 5 and a stride of 2. All maxpooling layers have a kernel size of 2 and stride of 2.}
  \label{fig:DepthArchi}
\end{figure}

The architecture of the model trained with depth maps is shown in Figure~\ref{fig:DepthArchi}, with the following parameters $F_1 = 5$, $F_2 = 5$, $F_3 = 5$, $F_4 = 10$. Note that the depth map has only one channel.

The model can be divided into the perception and control part. The perception part comprises of a series of CNN layers which extract the depth features of the depth maps (layers 1 to 8). These depth features are then flattened (layer 9) and passed through the depth control module (layer 10 to 12). After training is done, we replace the depth control module and concatenate the depth feature vector (of size $F_3 = 5$) with the segmentation embedding (of size $D=64$) and pass it through a sensor fusion control module for training. Note that while training this newly fused sensor control module the weights of the depth perception module are frozen. 

To train the model to also work when either one of the sensor fails we feed blank images for the depth sensor 30\% of the time while training. For the RGB images, there are two conditions of failure: 1) Like in the case of depth maps, the sensor fails to output RGB images, or 2) The sensor produces RGB images but the segmentation fails since the weather condition is different for which the segmentation training was done. For example, if we train our segmentation model on data from a clear afternoon, the model may fail if tested on a completely different scenario such as a rainy sunset. This second scenario, cannot be addressed by just simply feeding blank images. Instead, we randomly select an image from the training set that does not correspond to the current sample and feed it to the network, thus producing an incorrect segmentation embedding for the current time step. This is also done for 30\% of the training samples. For the remaining 40\% of the samples, both the correct depth maps and RGB images are fed simultaneously. Note that at each training instance, at least one of the two sensors always produces the correct information for the input to the control model.

\begin{figure}[ht]
  \centering
  \includegraphics[width=\linewidth]{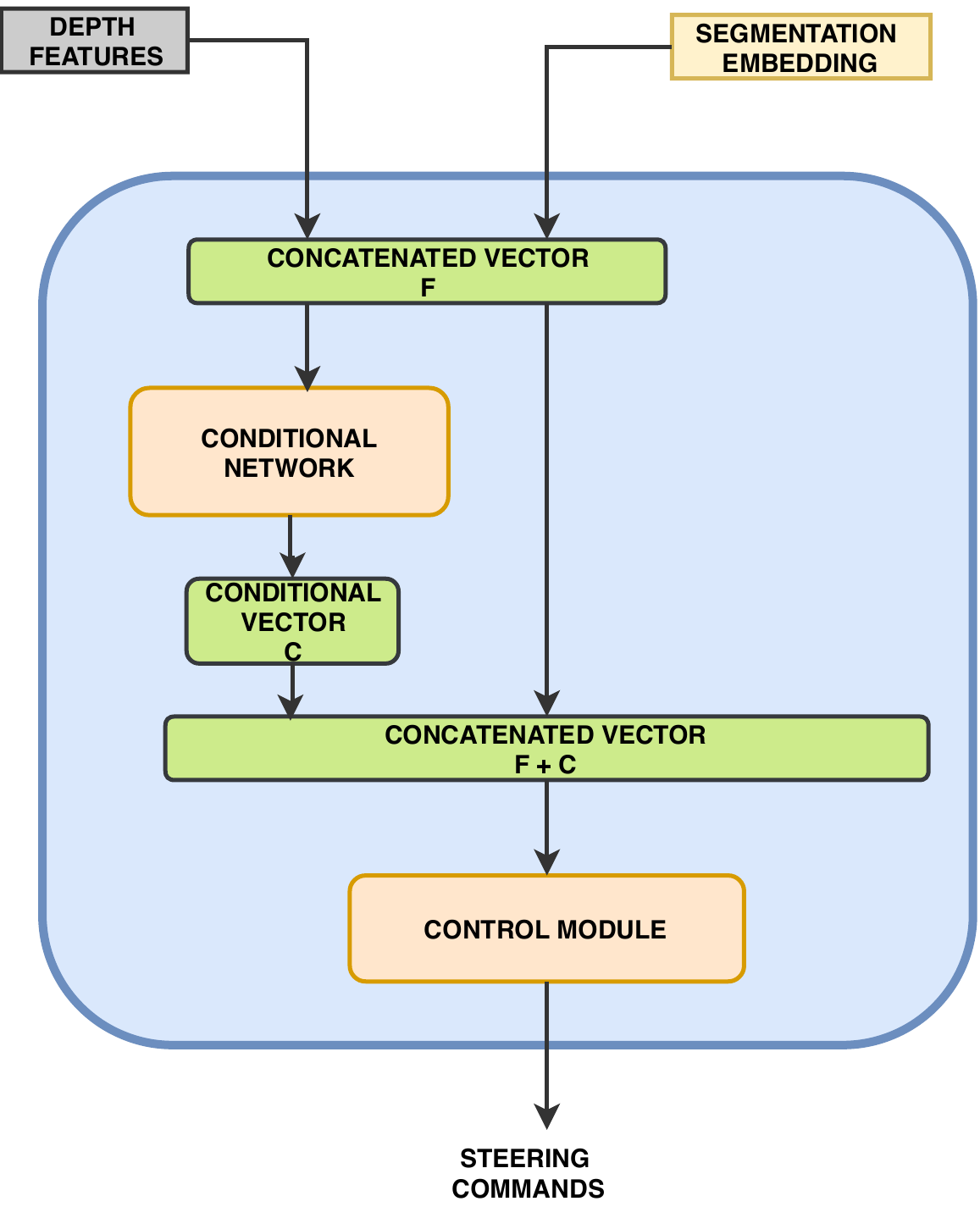}
  \caption{This figure shows the architecture for training a sensor fusion model. The depth features and the segmentation embeddings are concatenated into a new vector denoted by $F$. The vector $F$ is passed through a conditional network which outputs a conditional vector denoted by $C$. The conditional vector $C$ is then further concatenated with $F$ and passed through the control module, hence, being used for the prediction of the steering commands.}
  \label{fig:fusionmodel}
\end{figure}

\begin{figure}[ht]
  \centering
  \includegraphics[width=\linewidth]{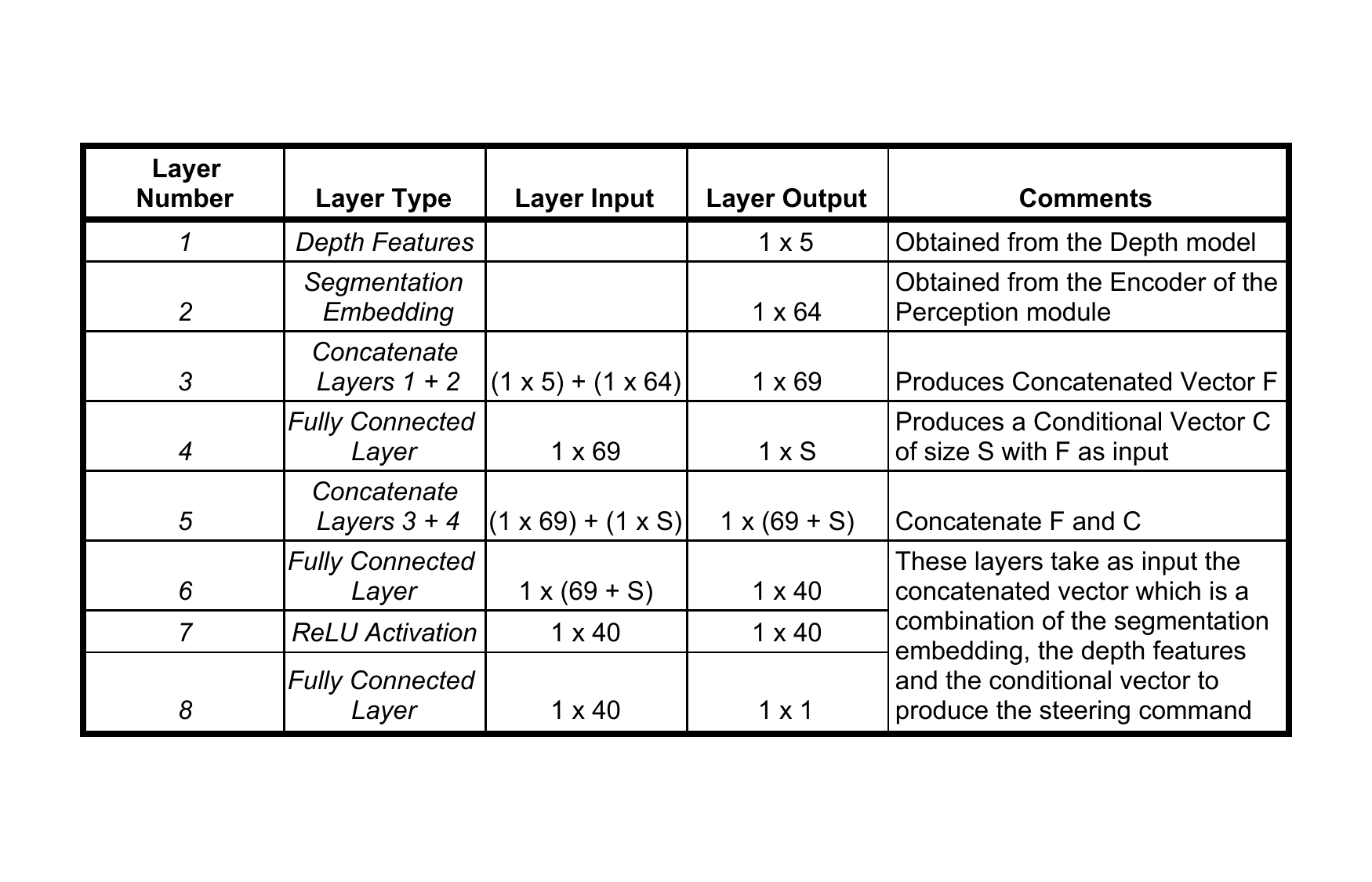}
  \caption{The architecture for training a control module by fusing information from the segmentation embedding and the depth features. First, a conditional vector of size $S$ is produced (layer 4 output) containing the inherent status of the 2 sensors. This vector is then concatenated with the depth features and segmentation embedding to ultimately furnish the control steering command. }
  \label{fig:SensorFusionArchi}
\end{figure}

\noindent{\textbf{Conditional network.}} The described method would not work by naively concatenating the depth feature vector with the segmentation embedding and directly feeding it to the control model. This is because failure of one sensor, may render information from the other functioning sensor to also be useless, since the control module was never trained to predict on one sensor only. Instead, the structure described in Figure~\ref{fig:fusionmodel} is used. We first pass the concatenated vector, $F$ through a conditional network. The purpose of this network is to produce a conditional vector $C$ which describes the inherent status of the two sensor modalities. Note that the conditional vector $C$ is learned in a self-supervised manner. Vector $C$ is then concatenated with vector $F$. This newly concatenated feature vector ($F + C$) is then passed through the control layers to predict the steering command of the vehicle. Acquainted with knowledge of the inherent status of the sensors contained in $C$, the control layers can appropriately assign relative importance to the inputs from the two senors while predicting the steering commands. The architecture of the sensor fusion control module is described in Figure~\ref{fig:SensorFusionArchi}.

With this approach we can train the conditional vector to ascertain the functioning status of the 2 sensors in a completely self-supervised manner. This approach is similar to one of the approaches proposed by~\cite{PatelIROS2017}. In that work instead of concatenating the vectors, the weighted sum of the 2 embeddings is taken. The 2 weights are multiplied with the relevant sensor embeddings before taking the sum. These weights are also learned by the model in a self-supervised manner. However, the approach of concatenation offers two apparent advantages over the weighted sum method:

\begin{itemize}
  \item To take the weighted sum, all the sensor embeddings must be of the same size. In our case the depth features are of size 5, whereas the segmentation embedding is of size 64. Therefore, before taking the weighted sum we would have had to add another layer to one of the sensor embeddings to bring it to the same size as that of the other sensor. However, with mere concatenation instead of taking weighted sum, we can evade ensuring that all sensor embeddings are of same size. 
  \item In our proposed method, the  conditional vector can be adjusted to any size depending on the complexity of the task and the nature of available sensors. Hence for complicated scenarios, we can increase the size of the conditional vector to embed more possibilities of representing the inherent status of all available sensors. This may not otherwise be possible by trying to represent the inherent status of a certain sensor by a single weight multiplier.
\end{itemize}

Nevertheless, with the weighted sum method, we can interpret the status of the sensors by observing the values of the weight multipliers. For a non-functioning sensor, the affiliated embedding would always have the same value corresponding to the failure condition (for \emph{e.g.} blank images). In that case we would expect its weight multiplier to also have a stable value. 

\begin{figure}[ht]
  \centering
  \includegraphics[width=\linewidth]{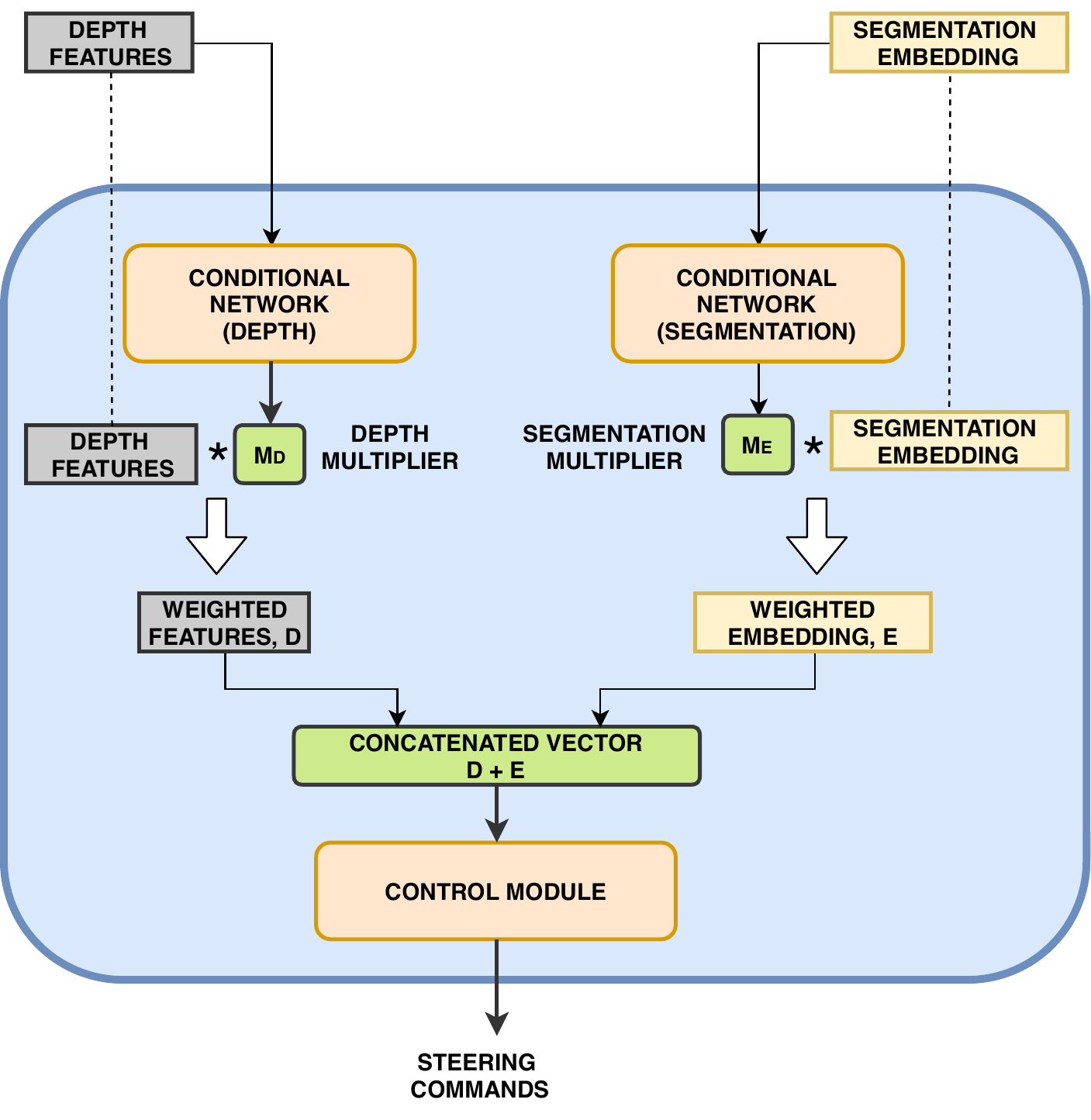}
  \caption{This figure shows the architecture for training a sensor fusion model with weight multipliers instead of a conditional vector.}
  \label{fig:CondVectorVisualizationModel}
\end{figure}

Therefore, to only understand and interpret the status of the 2 sensors, a different control model is trained which has weighted scalar multipliers for each sensor. However, instead of taking the weighted sum we take the weighted concatenation. Figure~\ref{fig:CondVectorVisualizationModel} shows the architecture for training a sensor fusion model with separate scalar weight multipliers instead of one joint conditional vector. The depth features and the segmentation embeddings are first passed through their respective conditional networks which produce the corresponding scalar weight multipliers $M_D$ (for depth features) and $M_E$ (for segmentation embedding). These scalars are multiplied with their affiliated sensor embeddings/features to produce the weighted depth feature $D$ and weighted segmentation embedding $E$. The feature vectors $D$ and $E$ are concatenated into a new vector which is passed through the control module to produce the steering command. Both the conditional networks have only one fully connected layer followed by a tanh activation function producing the scalar weight multiplier. The control module has the same architecture as that of layers 10 to 12 described in  Figure~\ref{fig:SensorFusionArchi}. The only difference is that the size of the concatenated vector input to the control module is of size 69 instead of $69 + S$. This is because we have eliminated the conditional vector of size $S$ by the weighted scalar multipliers.

\section{Experiments}

\begin{figure}[ht]
  \centering
  \includegraphics[width=\linewidth]{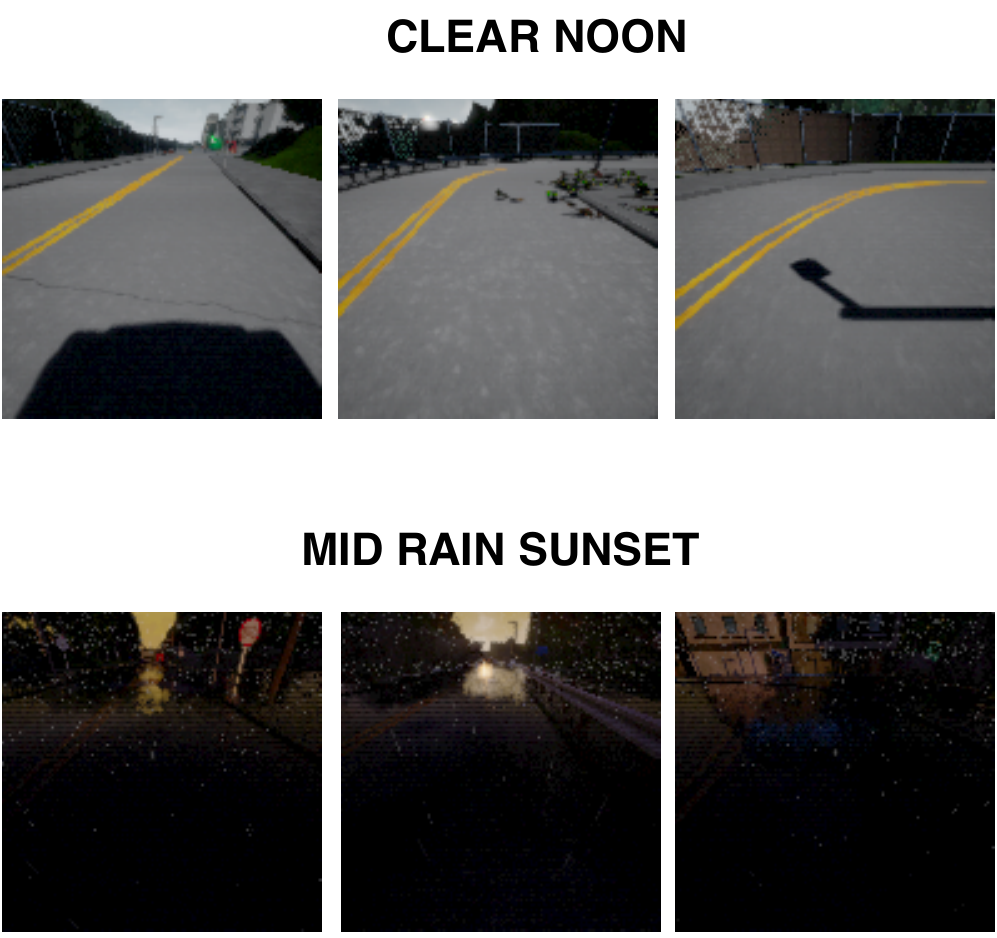}
  \caption{\textbf{Top:} 3 samples of the \emph{ClearNoon} weather condition on which the fusion model was trained. \textbf{Bottom:} 3 sample images of the \emph{MidRainSunset} weather condition.}
  \label{fig:weather_0_12_samples}
\end{figure}

To evaluate our approach we use the CARLA simulator~\cite{DosovitskiyCoRL2017}, which provides online feedback as a consequence of an ego-vehicle's steering angle prediction. We train the models with data collected (8000 samples) under the \emph{ClearNoon} weather condition but test under both the \emph{ClearNoon} and another different weather condition.  We chose the \emph{MidRainSunset}, as it is the very different from the default weather condition. Figure~\ref{fig:weather_0_12_samples} shows some samples of the two weather conditions, respectively. 

The collected samples correspond to RGB images and their semantic labels which are used to train a model to retrieve the semantic embedding. The simulator also produces the precise depth maps of the scene rendered using the Unreal engine \footnote{\url{https://www.unrealengine.com}}.

For these two weather conditions, we test the sensor fusion model under the following scenarios:

\begin{itemize}
  \item Input from both the RGB images and depth maps are available. However, note that the RGB image would not provide useful information in the \emph{MidRainSunset} condition because the segmentation would fail, since it was never trained for this weather condition. In that case we could expect the depth map to have a greater contribution in maneuvering the car in comparision with the RGB/semantic embedding.
  \item Input from only the depth sensor is available. In this case the RGB sensor input is zeroed out.
  \item Input from only the RGB sensor is available while the depth sensor input is zeroed out.
\end{itemize}

An additional test data set with 3000 samples for each weather condition was collected. Moreover, the starting positions of the samples are corresponding to right and left turns, respectively. This is due to the fact, that driving in a curve is a more complicated maneuver to accomplish in comparison with lane following. Hence, we can get a better understanding of sensor failure conditions. The size $S$ of the conditional vector in the fused control module described in Figure~\ref{fig:SensorFusionArchi} is chosen to be equal to 1. Furthermore, we use the segmentation model trained with the remapped labels as described in~\cite{KhanArXiv2019}.

Table~\ref{tab:fusionerrors} shows the Huber loss between the predicted and actual steering commands for the two different weather conditions under the three tested sensor configurations. We see that under the \emph{ClearNoon} weather condition, the error is the lowest when both sensors are functioning properly. When only either one of the sensors is functioning the errors are closely equivalent but still greater than the error for when both sensors are working. When tested live on the simulator, the model is capable of correctly making the turning maneuvers under all three sensor configurations for the \emph{ClearNoon} weather condition since it had been trained for this environment. Table~\ref{tab:videos_table} gives a description of the videos recorded, to visualize the online performance of the car on the CARLA simulator. Videos numbered 1, 2, and 3 describe the how the model performs under the 3 sensor configurations for the default weather condition.

\begin{table}[ht]
    \centering
    \caption{This table describes the Huber loss (in $10^{-3}$) for the two weather conditions under the three sensor configurations. The error is a dimensionless quantity giving the Huber loss between the predicted steering value from the fusion control model and the actual steering value. The CARLA steering values lie between $-1$ and $1$. The bold values represent the best performance for the respective weather condition.}
    \begin{tabular}{|c|c|c|c|}
        \hline
        Functioning Sensors & \emph{ClearNoon} & \emph{MidRainSunset}  \\
        \hline
        Both Sensors & \textbf{7.8} & 19.1\\
        \hline
        Depth only & 13.4 & \textbf{13.8} \\
        \hline
        RGB only & 15.7 & 32.7 \\
        \hline
    \end{tabular}
    \label{tab:fusionerrors}
\end{table}

For the \emph{MidRainSunset} weather condition, we see that the error when both sensors are functioning is higher than when both sensors are working under the \emph{ClearNoon} weather condition. This is expected, since the fusion model was never trained for \emph{MidRainSunset}. However, an interesting point to note is that the error is lower when only the depth sensor is functioning as compared to when both are working. This seems to imply that the RGB input is making it difficult for the fusion model to predict the correct steering command. 

\begin{figure}[ht]
  \centering
  \includegraphics[width=\linewidth]{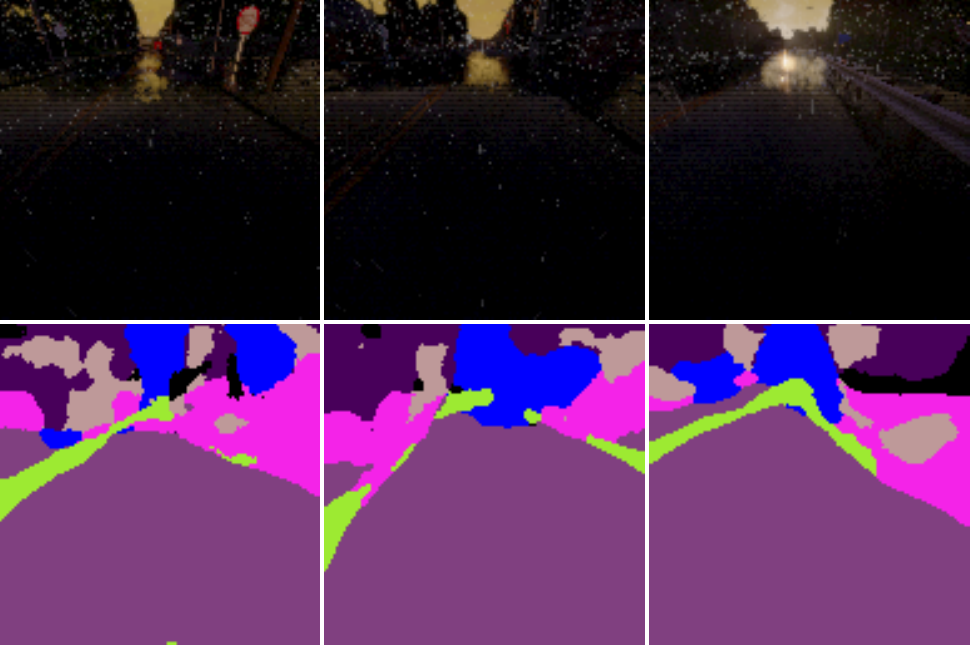}
  \caption{\textbf{Top:} 3 sample RGB images from the \emph{MidRainSunset} weather condition. \textbf{Bottom:} Corresponding reconstructed segmentation image from the segmentation embedding. The reconstruction is poor since the segmentation model was never trained for this weather condition.}
  \label{fig:segproblemsMRsunset}
\end{figure}

Figure~\ref{fig:segproblemsMRsunset} shows the reconstructed segmentation images for a sample of the three images under the \emph{MidRainSunset} condition. 

We see that since the segmentation model was never trained for this weather condition, the reconstruction is poor and in fact is adversely contributing towards the steering command prediction. The error is the highest when only input from the RGB sensor is considered. When tested live on the simulator, the RGB only configuration failed to perform as desired. Videos numbered 4, 5, and 6 in Table~\ref{tab:videos_table} describe how the model performs under the 3 sensor configurations for the \emph{MidRainSunset} condition.

\begin{figure}
  \centering
  \includegraphics[width=\linewidth]{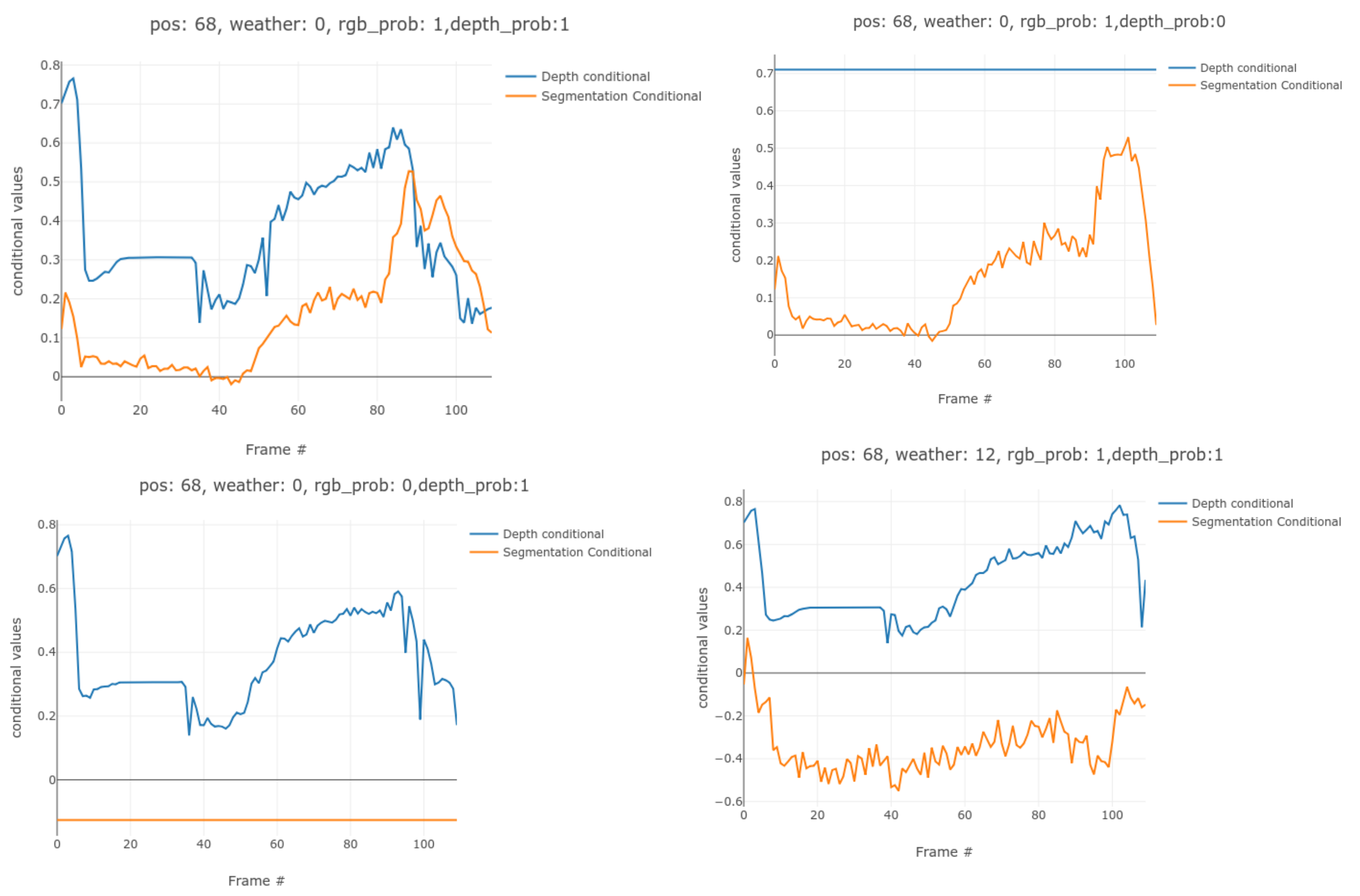}
  \caption{This figure shows 4 graphs for various configuration of sensors for the same starting position corresponding to a left turn. The turn is executed between frame 50-100. \textbf{Top left:} Weather: \emph{ClearNoon}, with both sensors. \textbf{Top right:} Weather: \emph{ClearNoon}, Depth map only, \textbf{Bottom left:} Weather: \emph{ClearNoon}, RGB image only, \textbf{Bottom right:} Weather: \emph{MidRainSunset}, both sensors. Note for weather \emph{ClearNoon} the conditional scalar of the functioning sensors fluctuates to a large degree while executing the turn. For weather \emph{MidRainSunset}, where visibility is low for the RGB camera its conditional scalar weight multiplier does not change much in comparison with that of the depth sensor.}
  \label{fig:CondVectorVisualization}
\end{figure}

\noindent{\textbf{Conditional vector visualization.}} The conditional vector of size $S$ produced by the fused control model described by the architecture in Figure~\ref{fig:SensorFusionArchi} is not interpretable for determining the inherent status of the various sensors. Therefore, we use the model described by Figure~\ref{fig:CondVectorVisualizationModel} to perform some additional tests to visualize the output of the conditional scalar weight multipliers of the 2 sensors. Figure~\ref{fig:CondVectorVisualization} shows some plots of the conditional scalar multipliers for the the depth and RGB sensors.

For a comparative analysis, all plots are for the same starting position corresponding to a left turn tested with different possibilities of sensor functioning on the \emph{ClearNoon} and \emph{MidRainSunset} weather condition. Videos numbered 7, 8, and 9 in Table~\ref{tab:videos_table} describes how the model performs under the 3 sensor configurations for the \emph{ClearNoon} condition using the weight multiplier method. Note in the videos that the frames corresponding to precisely when the car is actually executing the turn are from 50-100. Therefore, we draw our conclusions by observing the conditional values within these frames. It can be observed that on the \emph{ClearNoon} weather the functioning sensors show wide fluctuations between frames 50-100. The non-functioning sensors have a fixed value corresponding to the constant latent feature vector produced by the failure state of the sensor. It is interesting to note that for the \emph{MidRainSunset} condition, which has low visibility, the conditional vector for the RGB sensor remains fairly constant in comparison to that of the depth map, since the latent semantic vector is is not reliable enough to able to make a meaningful decision for controlling the car. 

\section{Conclusion}

From the experiments it can be observed that the car control module is capable of utilizing information from both the RGB and depth mapss to predict the correct steering commands. For the \emph{ClearNoon} weather condition, under which it was trained, the fused model was still capable of correctly maneuvering the car even when one of the 2 sensors failed. However, under the \emph{MidRainSunset} condition, it failed to give correct predictions when only the RGB camera was functioning, since the latent semantic embedding did not correspond to the correct reconstruction of the semantic map. In fact the performance of the fused model under the \emph{MidRainSunset} condition is better when only the depth sensor is functioning as opposed to when input from both sensors is received. 

\begin{sidewaystable}[ht]
\caption{This table describes the contents and performance of the videos which can be found at \href{https://www.youtube.com/playlist?list=PLKWxSGEZd0Ad8Iq-1ROQRoKNQaXt0xOMv}{https://www.youtube.com/playlist?list=PLKWxSGEZd0Ad8Iq-1ROQRoKNQaXt0xOMv.}}
\resizebox{\linewidth}{!}{
\begin{tabular}{|l||l|l|}
\hline
Video Id & Weather Type & Comments \\
\hline\hline
video1 & \emph{ClearNoon} & Tested with both the RGB and depth images fed to the network. The car successfully makes the turn since it was trained under this condition with both sensors functioning. \\
video2 & \emph{ClearNoon} & Only the input from the Depth image is fed to the network for the default weather condition. Meanwhile the RGB camera input is blanked out resulting in a segmentation output which is fixed. The car is still able to make the turn. \\
video3 & \emph{ClearNoon} & Only the input from the RGB image is fed to the network for the default weather condition. Meanwhile the Depth camera input is blanked out resulting in a segmentation output which is fixed. The car is still able to make the turn. \\
video4 & \emph{MidRainSunset} & Both RGB and Depth images are fed to the model. Since the visibility is low, the RGB camera is incapable of correctly reconstructing the semantic map. Nevertheless, the car is still capable of successfully maneuvering around the turn using information from the depth sensor. \\
video5 & \emph{MidRainSunset} & Only the Depth image is fed to the model, while the RGB image input is blanked out as can be seen from the fixed semantic reconstruction The car is  capable of successfully maneuvering around the turn using only the information from the depth sensor. \\
video6 & \emph{MidRainSunset} & Only the RGB image is fed to the model, while the Depth image input is blanked out. Due to low visibility the semantic reconstruction is poor and hence the car moves away from the correct trajectory and eventually crashes into the sidewalk. \\
video7 & \emph{ClearNoon} & Tested using the weighted multiplier approach. Both the RGB and Depth images are fed as input to the model. \\
video8 & \emph{ClearNoon} & Tested using the weighted multiplier approach. Just the depth image is fed to the network. As can be observed, only the weighted multiplier corresponding to the depth feature changes, whereas the multiplier for segmentation embedding is fixed.  \\
video9 & \emph{ClearNoon} & Tested using the weighted multiplier approach. Just the RGB image is fed to the network. As can be observed, only the weighted multiplier corresponding to the semantic embedding changes, whereas the multiplier for depth features is fixed.\\
\hline
\end{tabular}
}
\label{tab:videos_table}
\end{sidewaystable}

\bibliographystyle{IEEEtran}
\bibliography{main}

\end{document}